\pdfoutput=1
\documentclass{article}
\usepackage{custom}
\usepackage{xcolor}
\usepackage[section,below]{placeins}
\usepackage{float}
\usepackage{tabularx}
\usepackage{listings}
\lstdefinestyle{prompt}{basicstyle=\ttfamily\scriptsize, breaklines=true, breakindent=0pt, columns=fullflexible, keepspaces=true, breakatwhitespace=true, upquote=true, frame=single, rulecolor=\color{gray!50}, framesep=4pt, xleftmargin=4pt, xrightmargin=4pt, aboveskip=6pt, belowskip=8pt, escapeinside={(*@}{@*)}}

\title{Can LLMs predict the outcomes of machine learning experiments?}
\date{October 2026}
\author{%
  Ayush Bansal\textsuperscript{1} \quad
  Aldo Pareja\textsuperscript{2} \quad
  Sean Klein\textsuperscript{1} \quad
  Serge Didenko\textsuperscript{1} \quad
  Artem Lukoianov\textsuperscript{1,3} \\[0.6em]
  \textsuperscript{1}Autolab AI \quad
  \textsuperscript{2}Institute of Foundation Models \quad
  \textsuperscript{3}MIT
}

\begin{document}

\maketitle

\begin{abstract}
\hyphenpenalty=10000\exhyphenpenalty=10000
Recent advances in AI make Recursive Self-Improvement (RSI) increasingly plausible.
    Large Language Models (LLMs) can already propose, select, and run Machine Learning (ML) experiments.
    However, the space of possible design choices in ML is vast.
    An LLM that proposes an experiment needs a reliable intuition of whether it is worth running given the timing constraints of long training runs.
    In this work, we measure such intuition in current frontier LLMs by evaluating how accurately they predict the impact of self-generated ML experiment ideas before these are put into training runs.
    To this end, we construct a dataset of $288$ LLM-proposed ideas on eight MLE-bench tasks.
    We then prompt five frontier LLMs (GPT-6 Astra, Claude Opus 5.5, Claude Fable 5.1, Kimi K3, and GLM-5.3) to predict the impact of each idea and compare their predictions with the actual outcomes after training.
    We report four main findings.
    First, LLM predictions are substantially better than random chance but far from accurate: even the best model, Claude Opus 5.5, explains only half of the variance in outcomes ($R^2 = 0.49$).
    Second, we notice that all tested models systematically underestimate the magnitude of the changes.
    Third, prediction quality varies widely across tasks.
    Finally, the choice of the coding harness that actually implements the ideas has little effect on outcomes, and we find no evidence of self-preference.
    Higher reasoning settings and additional context lead to modest gains. Overall, our results show that current LLMs have useful but limited intuition about which ML ideas will work. We release our dataset to support further study.

\end{abstract}
\vfill
\begin{figure}[H]
    \centering
    \includegraphics[width=\linewidth]{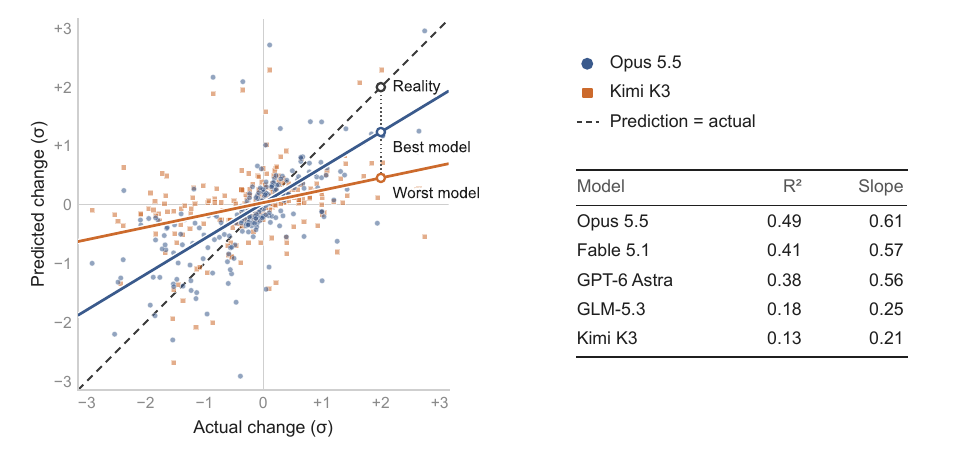}
    \caption{We prompt LLMs to predict how much an ML experiment idea will change the final training outcome. Left: predicted versus actual change for each idea, for the best model (Opus 5.5) and the worst model (Kimi K3), with each model's line of best fit. A perfect predictor would lie on the dashed line. Points beyond $\pm3\sigma$ are not shown. Right: $R^2$ and slope for all five models. Even the best model explains only half of the variance, and every fitted line is flatter than the diagonal, so all models underestimate large changes.}
    \label{fig:main}
\end{figure}
\vfill
\clearpage
\section{Introduction}
AI agents are increasingly used to run machine-learning experiments end to end~\cite{lu2024aiscientist,jiang2025aide,toledo2025aira}.
They are able to propose an idea, implement it, set up the training runs, evaluate the results, and propose a further improvement strategy.
This loop is often called autoresearch~\cite{karpathy2026autoresearch}, and is closely related to LLM-guided evolutionary search over programs~\cite{lehman2022elm,romera2024funsearch,novikov2025alphaevolve}.
While proposing new ideas is now cheap and fast, the training process itself is slow, which creates pressure to select only the most valuable ideas to balance out the high opportunity costs at training time.
In practice, many proposed ideas do not improve the score, so a large share of time and compute is spent on experiments that are later discarded.
To propose good ideas in the first place, an LLM needs an intuition of which ideas will work and which will not. Even when the model is never asked to judge its ideas explicitly, its early choices shape downstream experiments that it will later run, which compounds the cost of experimental runs.
This raises the question of whether LLMs can predict the outcome of their ideas before running them.

Prior works have evaluated the choices that an LLM makes when selecting the better of two ideas or solutions before running them~\cite{zheng2026foreagent,wen2025predicting,mule2026teaching,ji2026rehearse}, and found that LLMs pick the better option more often than chance~\cite{zheng2026foreagent,ji2026rehearse}.
Others asked for a predicted score, but used it only as a training reward~\cite{yang2026wmrl} or reported only an aggregate error~\cite{yang2026dsworld}.
Benchmarks such as MLE-bench~\cite{chan2024mlebench} evaluate how well agents solve ML tasks, not how well they anticipate the outcome of an idea.
A choice between two ideas only tells us which option a given LLM prefers, but not the magnitude of this preference.
In this work, we measure this magnitude by prompting a model for a numerical estimate of the score after the idea is implemented.
We build a dataset of $288$ ML ideas with measured outcomes, prompt five frontier LLMs to predict the score after each idea, and measure how closely the predictions approximate the actual results after training.

In Section~\ref{sec:related}, we review related work. In Section~\ref{sec:method}, we describe how we build ML experiments with known outcomes and collect each model's prediction. In Section~\ref{sec:results}, we compare these predictions to the real results. In Section~\ref{sec:ablations}, we test whether a higher reasoning level or more context in the prompt helps, and then whether the model implementing an idea matters and whether models have self-preference. 

Our contributions are as follows:
\begin{enumerate}
    \item We propose a way to measure the ML intuition of LLMs, and release a dataset of $288$ LLM-proposed ideas on eight MLE-bench tasks together with the real change each one produced.\footnote{\url{https://github.com/autolab-ai/EV-estimation}}
    \item We evaluate how well five current frontier LLMs (GPT-6 Astra, Claude Opus 5.5, Claude Fable 5.1, Kimi K3, and GLM-5.3) estimate the effect of ML engineering ideas.
    \item We vary the reasoning level and the context provided to the predictor, ablate the implementing model, and test for self-preference.
\end{enumerate}

\section{Related work}
\label{sec:related}

\paragraph{ML research agents.}
Prior work on ML research agents has mostly focused on building better agents, such as The AI Scientist~\cite{lu2024aiscientist,lu2026automation}, AIDE~\cite{jiang2025aide}, AIRA~\cite{toledo2025aira}, and autoresearch~\cite{karpathy2026autoresearch}. These agents are usually evaluated on benchmarks of ML tasks, such as MLE-bench~\cite{chan2024mlebench}, RE-Bench~\cite{wijk2024rebench}, MLAgentBench~\cite{huang2024mlagentbench}, MLGym~\cite{nathani2025mlgym}, and PaperBench~\cite{starace2025paperbench}, which score an agent by the results its code achieves. Most agents decide which ideas to keep by running every candidate. However, some use an LLM's judgment to decide which ideas to run in full~\cite{weng2025deepscientist,kulibaba2025kompeteai,choi2026aim}, while others avoid it altogether, since LLM self-assessments are often unreliable~\cite{yu2026discovery}. The quality of this judgment therefore directly affects how efficiently such agents work.

\paragraph{Predicting the outcome of an experiment.}
Closest to our work are studies that ask LLMs to pick the better of two ideas or solutions before running them, either on MLE-bench solutions~\cite{zheng2026foreagent}, on research ideas~\cite{wen2025predicting,mule2026teaching}, or inside agent loops~\cite{ji2026rehearse,foster2026preference}. For ML solutions, LLMs pick the better option more often than chance, but far from perfectly~\cite{zheng2026foreagent,foster2026preference}. For research ideas, the best results come from models fine-tuned on past outcomes~\cite{wen2025predicting,mule2026teaching}. Other works ask for a predicted score instead. In particular, DSWorld~\cite{yang2026dsworld} and WMRL~\cite{yang2026wmrl} use predicted scores in place of real runs, and LLMs have also been used to predict the performance of hyperparameters, architectures, and GPU kernels~\cite{liu2024llambo,jawahar2024llmpp,khan2026gpuforecasters}. Concurrent with our work, Wang et al.~\cite{wang2026rwm} ask LLMs to predict the gain of code and configuration changes in LLM research environments and show that records of past experiments improve these predictions. We focus on the intuition a model brings to a new task before it has any such records. We also measure not only how well the predictions rank the ideas, but also whether their magnitude and direction match the actual changes.

\paragraph{Forecasting with LLMs.}
More broadly, LLMs are increasingly used to forecast real-world events~\cite{halawi2024forecasting,karger2024forecastbench,alur2025aia} and the results of experiments in other sciences~\cite{ashokkumar2026social}. These forecasts often have systematic biases. For example, LLMs hedge their probabilities toward $50\%$ on binary questions~\cite{alur2025aia}, while they overestimate the size of effects in social science experiments~\cite{ashokkumar2026social}. In this work, we study the same question for ML experiments.

\section{Methodology}
\label{sec:method}

\subsection{Dataset}
We build a dataset of ML ideas, each paired with the measured change in score that the implemented idea produced. We start from a baseline solution and its score, and prompt an LLM to propose an idea intended to either improve or worsen the score. The same model then implements the idea and trains the resulting model, which we evaluate to obtain the new score. For example, one of our baselines uses TF-IDF features with logistic regression and has a log loss of $0.432$. Fable 5.1 proposed adding a second model and averaging its predictions with those of the first, which reduced the log loss to $0.361$.

\begin{figure}[!t]
    \centering
    \includegraphics[width=\linewidth]{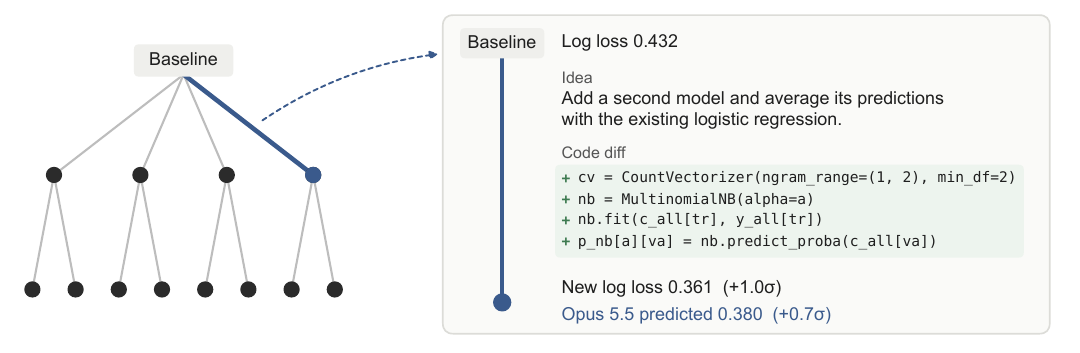}
    \caption{Left: starting from the baseline, a model proposes four ideas and then two follow-up ideas for each. Each node is a working solution and each edge is an idea. Right: a real example from our dataset, showing the idea, its code diff, and the predicted versus actual score after training. Here, Opus 5.5 predicts the right direction but underestimates the size of the improvement.}
    \label{fig:setup}
\end{figure}

Repeating this process from the baseline and from each resulting solution produces a tree, as shown in Figure~\ref{fig:setup}. Each idea is an edge, and the solution it leads to is a node. The model proposes four ideas for the baseline, and then two follow-up ideas for each of the four new solutions, for a total of $12$ ideas per tree.

We collect ideas on eight ML tasks from MLE-bench Lite~\cite{chan2024mlebench}, which cover three kinds of metrics: log loss (spooky author identification, dog breed identification, and leaf classification), ROC AUC (tabular playground series May 2022, toxic comment classification, and random acts of pizza), and RMSE (New York City taxi fares and denoising dirty documents). All eight are past Kaggle competitions with public leaderboards. For each, we choose a baseline that scores near the middle of the leaderboard, which leaves room for ideas to both improve and worsen the score.

For each task, we repeat this process three times, once each with GPT-6 Astra, Fable 5.1, and Kimi K3 proposing and implementing the ideas. This yields $8$ tasks $\times$ $3$ models $\times$ $12$ ideas $= 288$ ideas. We exclude $19$ ideas whose runs failed or for which a model returned no usable prediction, leaving $269$ ideas for analysis.

\paragraph{Uniform sampling.}
Our goal is to measure how well LLMs estimate the outcome of an idea, whether that outcome is an improvement or a degradation of the score. To sample outcomes more evenly in both directions, we prompt the model to improve the score for half of the ideas in each tree and to worsen it for the other half. These are plausible, good-faith proposals rather than deliberately broken changes (Appendix~\ref{app:prompts}). In practice, many ideas still move the score in the unintended direction. Appendix~\ref{app:ideadirection} shows how often each model's ideas moved the score in the prompted direction, and Appendix~\ref{app:examples} shows example ideas.

\paragraph{Normalization.}
The eight tasks use different metrics on very different scales. For example, a change of $0.01$ means something very different for ROC AUC, which lies between $0$ and $1$, than for the RMSE of taxi fares, which is measured in dollars. To compare and pool results across tasks, we therefore measure every outcome in standard deviations of the outcomes on the same task. For each idea, we take the difference between the new score and the score of the solution the idea was applied to, and flip its sign for metrics where lower is better, so that a positive change always means an improvement. We then divide this change by $\sigma$, the standard deviation of the actual changes across all ideas that the three models ran on that task. We normalize each predicted change in the same way, using the predicted score and the same $\sigma$, so that predictions and outcomes are on one scale. Thus, a change of $+1\sigma$ is an improvement as large as the typical spread of outcomes on that task.

\subsection{Measuring ML intuition}
We define a model's intuition as how closely its predicted changes in score follow the measured outcomes. To measure this, we prompt five LLMs to predict the result of every change. We use Opus 5.5, Fable 5.1, GPT-6 Astra, Kimi K3, and GLM-5.3, all at their maximum reasoning level. For each change, the model is given the task, the current code with its score, the idea, and the code diff, and then predicts the new score (Appendix~\ref{app:prompts}). For each model, we then fit a least-squares line to its predicted changes as a function of the actual changes (Figure~\ref{fig:main}). We summarize the model by two properties of this line: its $R^2$, the share of the variation in the predictions that the actual outcomes explain, which tells us how well the model separates better ideas from worse ones (our primary measure); and its slope, which tells us whether the predicted changes have the right size (secondary measure).

We do not expect LLMs to be calibrated to output the exact score change for every metric of every ML task. Moreover, in autoresearch the exact value matters less than how the ideas compare to each other, since we want to pick the best one out of many candidates. We therefore focus on how closely the points follow the fitted line, which we measure with $R^2$. The slope is a matter of calibration, and the intercept is close to zero for every model. Rank correlation measures, such as Spearman's $\rho$, provide a reasonable alternative. However, a rank measure only uses the order of the ideas, like a choice between two ideas. We focus on $R^2$, since it also uses how much better one idea is expected to be than another, and in the regime of many candidate ideas it directly determines the expected quality of the selected idea.
Under a simple Gaussian model, the idea a model ranks highest gives on average $\sqrt{R^2}$ of the gain of the best idea, both measured from the average idea. Thus, a model with $R^2 = 1$ always finds the best idea, while a model with $R^2 = 0$ does no better than picking at random.

The slope remains informative, as it indicates whether a model anticipates the magnitude of a change. A slope of $1$ means the predictions match the actual size of each change, whereas a slope below $1$ means the model underestimates the largest changes. Magnitude matters for separating ideas that yield larger improvements, such as changes to the model architecture, from ideas that likely yield only small gains.

\section{Results}
\label{sec:results}
Figure~\ref{fig:main} shows each model's predicted changes against the actual ones, with the fitted lines. Opus 5.5 performs best ($R^2 = 0.49$), followed by Fable 5.1 and GPT-6 Astra at $R^2 = 0.41$ and $R^2 = 0.38$, respectively. These three models substantially outperform the open-weight models GLM-5.3 and Kimi K3, which reach only $R^2 = 0.18$ and $R^2 = 0.13$, respectively. Under the Gaussian model from Section~\ref{sec:method}, the idea ranked highest by Opus 5.5 achieves on average about $70\%$ of the gain of the best idea, compared with $64\%$ for Fable 5.1, $62\%$ for GPT-6 Astra, $42\%$ for GLM-5.3, and $36\%$ for Kimi K3.

However, every model has a slope below $1$, that is, every model underestimates how much an idea will change the score. Opus 5.5 has the highest slope at $0.61$, followed closely by Fable 5.1 and GPT-6 Astra at $0.57$ and $0.56$, respectively, while GLM-5.3 and Kimi K3 have much lower slopes at $0.25$ and $0.21$, respectively. For instance, an actual change of $+2\sigma$ corresponds to a predicted change of about $+1.2\sigma$ for the best model and $+0.5\sigma$ for the worst (Figure~\ref{fig:main}).

\begin{figure}[t]
    \centering
    \includegraphics[width=0.65\linewidth]{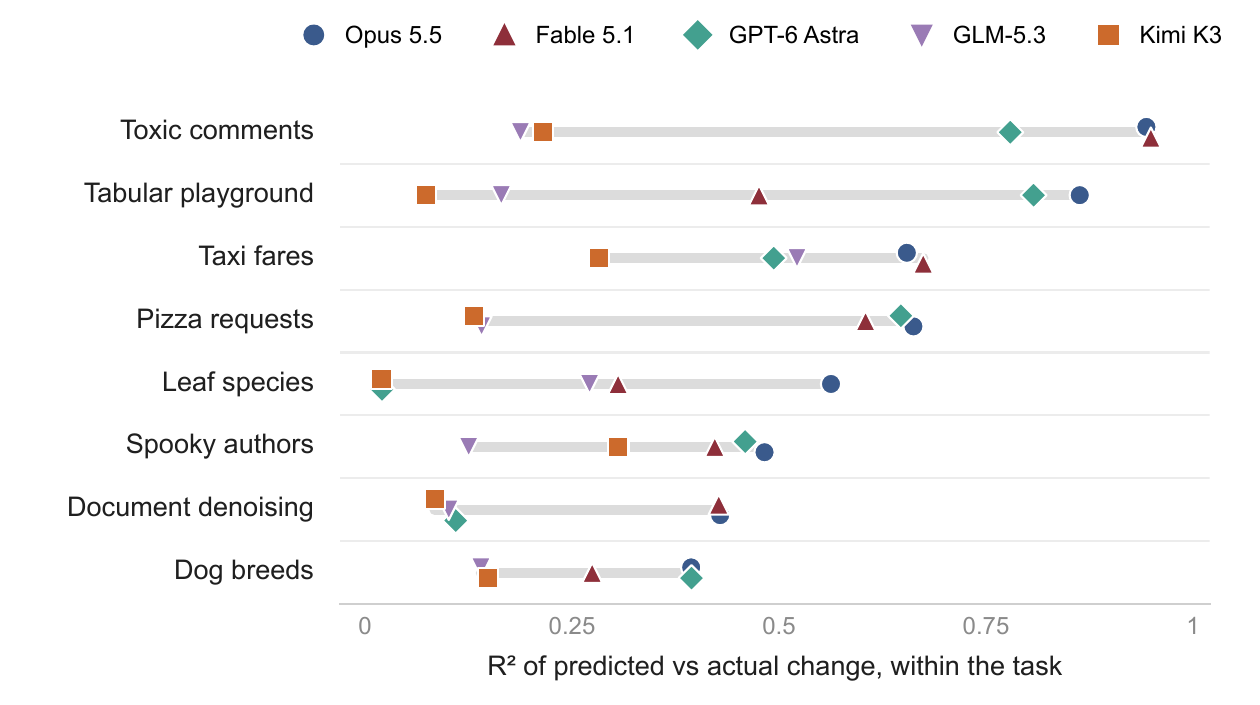}
    \caption{$R^2$ per task. Each row is one MLE-bench task and each marker one model. The same model can be highly accurate on one task and poor on another, and Opus 5.5 beats Kimi K3 on every task. The two image tasks, dog breeds and document denoising, are the hardest for the best model.}
    \label{fig:task}
\end{figure}

Figure~\ref{fig:main} pools all eight tasks. Computing $R^2$ separately for each task (Figure~\ref{fig:task}) reveals large differences: the same model can be highly accurate on one task and poor on another. For example, Opus 5.5 reaches $R^2 = 0.94$ on toxic comment classification but only $0.39$ on dog breed identification. Similarly, GPT-6 Astra ranges from $0.81$ on the tabular playground task to $0.02$ on leaf classification. The two image tasks, dog breed identification and denoising dirty documents, are the hardest to predict: even the best model reaches only $R^2 = 0.39$ and $0.43$.

\begin{figure}[!hb]
    \centering
    \includegraphics[width=0.9\linewidth]{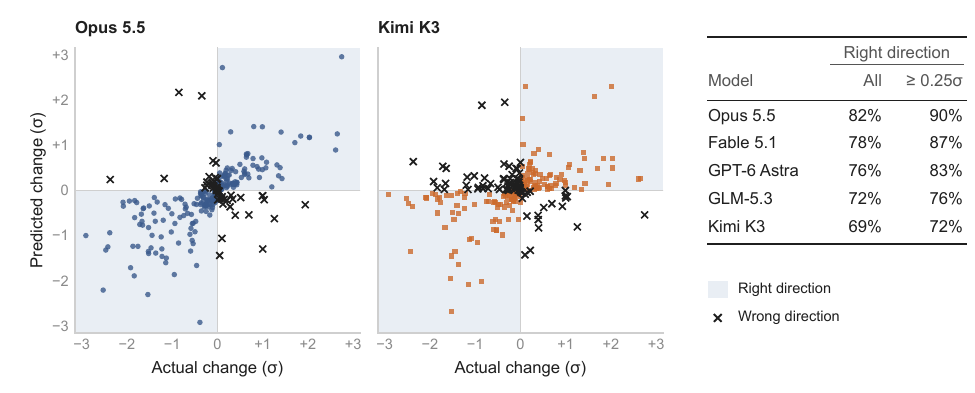}
    \caption{Each point is one idea, for the best model (Opus 5.5) and the worst model (Kimi K3). Points in the shaded quadrants have the correct predicted direction, and crosses mark ideas where the model predicted the wrong direction. Most ideas cluster near zero, so most have only a modest effect on the score. The table shows how often each model predicts the right direction, for all ideas and for ideas that changed the score by at least $0.25\sigma$. Points beyond $\pm3\sigma$ are not shown. Every model predicts the direction of most changes correctly, and accuracy rises once small changes are excluded.}
    \label{fig:direction_predict}
\end{figure}
The models are considerably better at predicting the direction of a change than its magnitude (Figure~\ref{fig:direction_predict}). Opus 5.5 predicts the correct direction for $82\%$ of the ideas, Fable 5.1 $78\%$, GPT-6 Astra $76\%$, GLM-5.3 $72\%$, and Kimi K3 $69\%$. However, many ideas lie between $-0.25\sigma$ and $+0.25\sigma$, where the sign of the change may depend on the particular training and test split rather than on the idea itself. Excluding these ideas increases every model's accuracy (table in Figure~\ref{fig:direction_predict}).

\section{Ablations}
\label{sec:ablations}
We now ablate the main components of our system. First, we change what the predictor gets, namely its reasoning level and the context in its prompt. Then, we check two possible confounds in how the dataset was built: the model that implements each idea, and whether models favor their own ideas.

\subsection{Reasoning level}
\begin{figure}[!htb]
    \centering
    \includegraphics[width=0.74\linewidth]{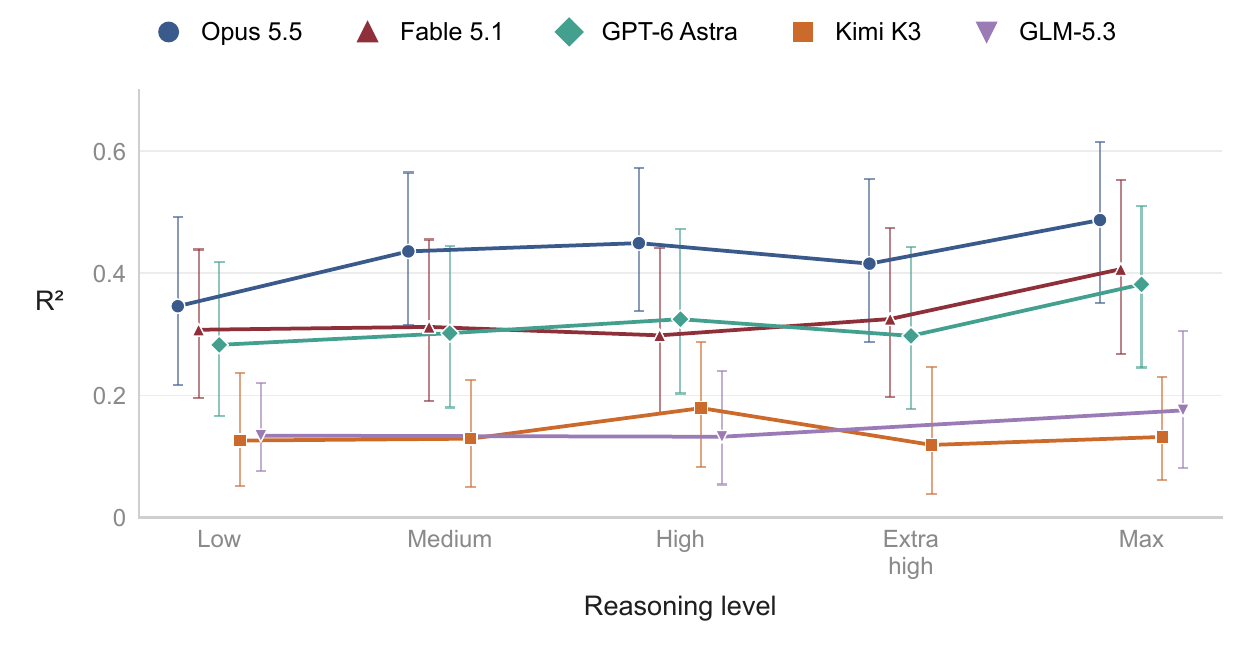}
    \caption{$R^2$ of each model as a function of reasoning level, from low to max. GLM-5.3 only has three reasoning levels, while all others have five. Error bars are $95\%$ intervals from bootstrapping over the ideas. $R^2$ rises with reasoning level for the three strongest models, but not for Kimi K3 and GLM-5.3.}
    \label{fig:reasoning}
\end{figure}
\begin{figure}[t]
    \centering
    \includegraphics[width=0.74\linewidth]{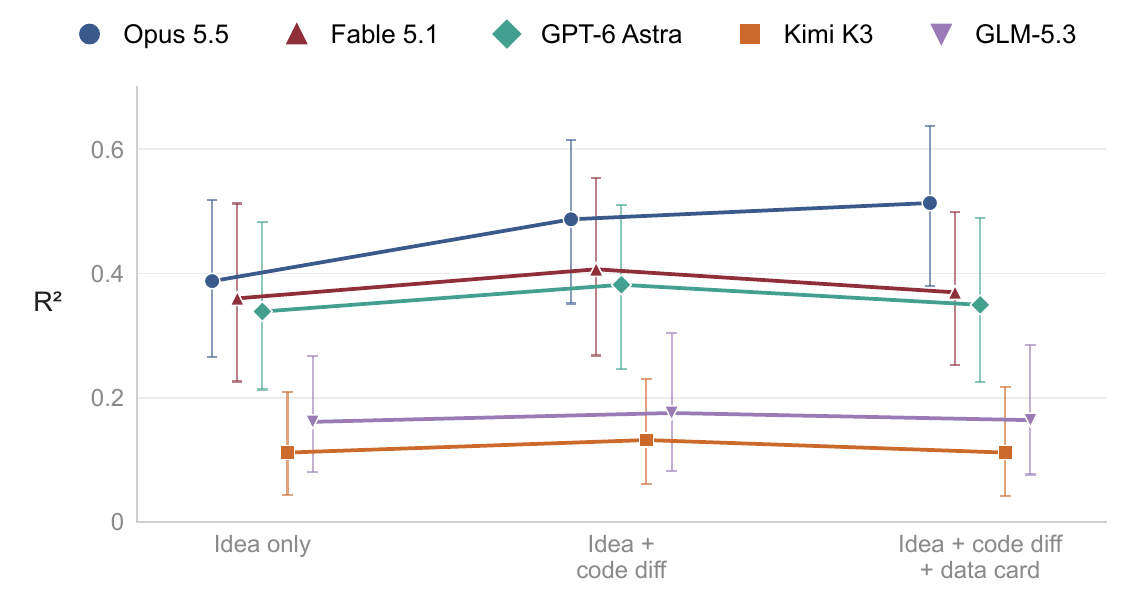}
    \caption{From left to right, the model sees only the idea, the idea and its code diff, or the idea, its code diff, a data card describing the data, and the first $20$ rows of the data. Error bars are $95\%$ intervals from bootstrapping over the ideas. Removing the code diff lowers $R^2$ for the three strongest models, while the data card helps only Opus 5.5.}
    \label{fig:context}
\end{figure}

All main results use each model at its maximum reasoning level. Here, we test whether additional reasoning improves the predictions. Prior work finds that longer reasoning helps primarily on math and symbolic problems~\cite{sprague2025cot}, and demonstrates no consistent gain on numeric estimates~\cite{hobor2026bayesian}. We find that $R^2$ increases with the reasoning level for the three strongest models, with Opus 5.5 going from $0.35$ to $0.49$, Fable 5.1 from $0.31$ to $0.41$, and GPT-6 Astra from $0.28$ to $0.38$ (Figure~\ref{fig:reasoning}). For Kimi K3 and GLM-5.3, $R^2$ varies little across reasoning levels. Although the bootstrapped error bars overlap, the gain from the lowest to the highest reasoning level is clearly above zero for GPT-6 Astra and nearly so for Opus 5.5. Additional reasoning therefore helps only the strongest models.

\subsection{Context provided}
\label{sec:context}
Next, we test whether a model predicts better when it has more information. In Figure~\ref{fig:direction_predict}, many ideas cluster between $-0.25\sigma$ and $+0.25\sigma$, and for these, whether an idea helps can depend on the training data itself. To test this, we add to the prompt a ``data card'' that describes the data, together with its first $20$ rows, since the full dataset does not fit in the context window. In a second test, we provide only the idea, without the code diff.
Removing the code diff degrades the three strongest models, whereas adding the data card helps only Opus 5.5 (Figure~\ref{fig:context}). Kimi K3 and GLM-5.3 are largely unaffected. However, most changes are within the error bars, and only Opus 5.5's overall gain from the idea alone to the full prompt is clearly larger than the noise. Overall, additional context benefits the strongest model but adds little for the others.

\subsection{Implementer of the ideas}
\begin{figure}[!htb]
    \centering
    \includegraphics[width=0.82\linewidth, trim=0 0 0 28bp, clip]{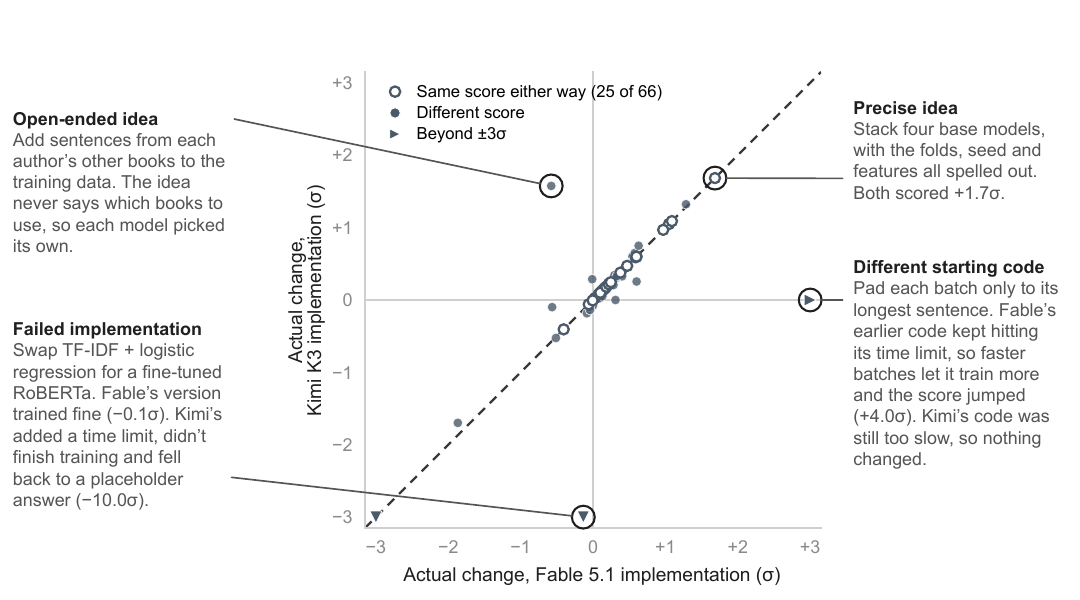}
    \caption{Each point is one idea, plotted at its change in score when implemented by Fable 5.1 (x-axis) and by Kimi K3 (y-axis). Points on the dashed line scored identically under both implementations. The four circled ideas are examples where the implementer does and does not matter. Most ideas score nearly the same regardless of the implementer, with $58$ of $66$ within $0.25\sigma$.}
    \label{fig:executor}
\end{figure}
In our main dataset, each idea is implemented by the same model that proposed it. Since an outcome depends on both the text description of the idea and its implementation, i.e.\ the exact code diff, predicting it is only well defined if different implementations of the same idea give similar results. To test this, we collect a separate set of ideas on three tasks, in which Kimi K3 and Fable 5.1 propose ideas and each model implements both its own ideas and those of the other. Most ideas yield nearly the same result under either implementation (Figure~\ref{fig:executor}), with $58$ of $66$ within $0.25\sigma$ of each other and $25$ giving the exact same score. Precise ideas that specify the exact settings produce the same result regardless of the implementer. The implementer changes the result substantially in only a few cases. Some ideas are open-ended and leave choices to the implementer, such as which additional data to use, so the two models effectively run different experiments. Others fail in one implementation, for example when a model imposes its own time limit, stops before training finishes, and falls back to a trivial prediction. Overall, the implementer has little effect.

\subsection{Self-preference}

\begin{figure}[t]
    \centering
    \includegraphics[width=0.45\linewidth]{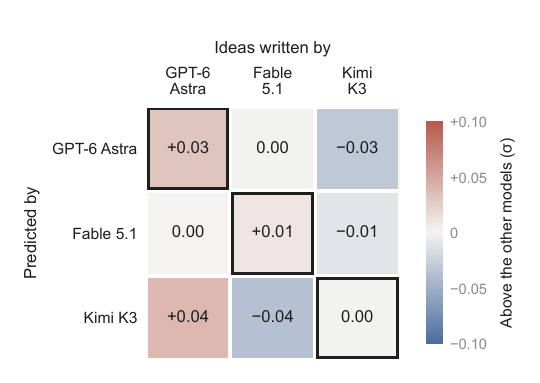}
    \caption{Each cell shows how far one model's predictions (rows) lie above its own line of best fit, relative to the other four models on the same ideas, averaged over the ideas written by each model (columns). The outlined diagonal shows each model predicting its own ideas, where self-preference would appear as positive values. All diagonal values are close to zero, so we find no self-preference.}
    \label{fig:selfpref}
\end{figure}

Since the same models both write ideas and predict them, a model could be more optimistic about its own ideas, a phenomenon known as self-preference. LLM judges have been found to rate their own outputs higher~\cite{zheng2023judging,panickssery2024llm}, although part of this goes away once the quality of the answers and of the judge is controlled for~\cite{chen2025prefer,roytburg2026narcissists}. To test this, we compute, for each idea, how far a model's prediction lies above that model's line of best fit (Figure~\ref{fig:main}). We then compare this residual with those of the other four models on the same idea, which controls for idea quality and each model's overall optimism. All three diagonal cells in Figure~\ref{fig:selfpref} are close to zero (at most $+0.03\sigma$). On average, a model's residual on its own ideas exceeds that of the other models by only $0.01\sigma$ ($95\%$ interval $-0.06\sigma$ to $+0.08\sigma$). We therefore find no evidence of self-preference.

\section{Conclusions}
\label{sec:conclusions}
In this work, we asked whether LLMs can predict the outcome of an ML idea before running it. To this end, we built a dataset of ML experiments with known outcomes and prompted five frontier models to predict how each idea would change the score. Even the best model, Opus 5.5, explains only about half of the variance in outcomes ($R^2 = 0.49$), while GLM-5.3 and Kimi K3 explain very little. All tested models underestimate the magnitude of large changes, and prediction quality depends strongly on the task. The models are more reliable at predicting the direction of a change, particularly for larger changes, than at identifying which ideas will have a large effect. Our ablations show that additional reasoning helps only the strongest models, that additional context helps only Opus 5.5, that the implementing model has little effect, and that there is no evidence of self-preference.

For autoresearch, these results suggest that model predictions can already help prioritize which ideas to run, but cannot yet replace running the experiments. This matters most for recursive self-improvement (RSI), where one experiment may mean training a new model and choosing the wrong ideas is costly.

Our study has several limitations. Each idea was run only once, so part of each measured change may reflect training noise, and we do not know the ceiling that a perfect predictor could reach. Moreover, we evaluate only eight tasks, all of them small ML problems rather than LLM training runs. Future work could rerun ideas with multiple seeds to quantify this noise, extend the evaluation to larger tasks and codebases, and test whether models can predict the outcome of a sequence of ideas rather than a single one.

\clearpage
\bibliographystyle{unsrt}
\bibliography{biblio}

\begin{thebibliography}{10}

\bibitem{lu2024aiscientist}
Chris Lu, Cong Lu, Robert~Tjarko Lange, Jakob Foerster, Jeff Clune, and David
  Ha.
\newblock The {AI} {S}cientist: Towards fully automated open-ended scientific
  discovery.
\newblock {\em arXiv preprint arXiv:2408.06292}, 2024.

\bibitem{jiang2025aide}
Zhengyao Jiang, Dominik Schmidt, Dhruv Srikanth, Dixing Xu, Ian Kaplan, Deniss
  Jacenko, and Yuxiang Wu.
\newblock {AIDE}: {AI}-driven exploration in the space of code.
\newblock {\em arXiv preprint arXiv:2502.13138}, 2025.

\bibitem{toledo2025aira}
Edan Toledo, Karen Hambardzumyan, Martin Josifoski, Rishi Hazra, Nicolas
  Baldwin, Alexis Audran-Reiss, Michael Kuchnik, Despoina Magka, et~al.
\newblock {AI} research agents for machine learning: Search, exploration, and
  generalization in {MLE}-bench.
\newblock In {\em Advances in Neural Information Processing Systems (NeurIPS)},
  volume~38, 2025.
\newblock arXiv:2507.02554.

\bibitem{karpathy2026autoresearch}
Andrej Karpathy.
\newblock autoresearch: {AI} agents running research on single-{GPU} nanochat
  training automatically.
\newblock \url{https://github.com/karpathy/autoresearch}, 2026.

\bibitem{lehman2022elm}
Joel Lehman, Jonathan Gordon, Shawn Jain, Kamal Ndousse, Cathy Yeh, and
  Kenneth~O. Stanley.
\newblock Evolution through large models.
\newblock In Wolfgang Banzhaf, Penousal Machado, and Mengjie Zhang, editors,
  {\em Handbook of Evolutionary Machine Learning}, pages 331--366. Springer
  Nature Singapore, Singapore, 2023.

\bibitem{romera2024funsearch}
Bernardino Romera-Paredes, Mohammadamin Barekatain, Alexander Novikov, Matej
  Balog, M.~Pawan Kumar, Emilien Dupont, Francisco J.~R. Ruiz, Jordan~S.
  Ellenberg, Pengming Wang, Omar Fawzi, Pushmeet Kohli, and Alhussein Fawzi.
\newblock Mathematical discoveries from program search with large language
  models.
\newblock {\em Nature}, 625:468--475, 2024.

\bibitem{novikov2025alphaevolve}
Alexander Novikov, Ng{\^a}n V{\~u}, Marvin Eisenberger, Emilien Dupont, Po-Sen
  Huang, Adam~Zsolt Wagner, Sergey Shirobokov, Borislav Kozlovskii, Francisco
  J.~R. Ruiz, Abbas Mehrabian, M.~Pawan Kumar, Abigail See, Swarat Chaudhuri,
  George Holland, Alex Davies, Sebastian Nowozin, Pushmeet Kohli, and Matej
  Balog.
\newblock {AlphaEvolve}: A coding agent for scientific and algorithmic
  discovery.
\newblock {\em arXiv preprint arXiv:2506.13131}, 2025.

\bibitem{zheng2026foreagent}
Jingsheng Zheng, Jintian Zhang, Yujie Luo, Yuren Mao, Yunjun Gao, Lun Du,
  Huajun Chen, and Ningyu Zhang.
\newblock Can we predict before executing machine learning agents?
\newblock In {\em Proceedings of the 64th Annual Meeting of the Association for
  Computational Linguistics (Volume 1: Long Papers)}, pages 3941--3974, 2026.
\newblock arXiv:2601.05930.

\bibitem{wen2025predicting}
Jiaxin Wen, Chenglei Si, Yueh-Han Chen, He~He, and Shi Feng.
\newblock Predicting empirical {AI} research outcomes with language models.
\newblock In {\em Advances in Neural Information Processing Systems (NeurIPS)},
  2025.
\newblock arXiv:2506.00794.

\bibitem{mule2026teaching}
Srujan~P. Mule, Aniketh Garikaparthi, and Manasi Patwardhan.
\newblock Teaching language models to forecast research success through
  comparative idea evaluation.
\newblock In {\em Findings of the Association for Computational Linguistics:
  ACL 2026}, 2026.
\newblock arXiv:2605.21491.

\bibitem{ji2026rehearse}
Jiazhen Ji and Shouhong Ding.
\newblock Rehearse: Stepping back from the confidence cliff in self-improving
  autoresearch.
\newblock {\em arXiv preprint arXiv:2607.27687}, 2026.

\bibitem{yang2026wmrl}
Xiyuan Yang, Sheikh Sarwar, Jingru Cheng, Zhan Shi, Duanshun Li, Huiyuan Chen,
  Haiyang Zhang, Xing Fan, Chenlei Guo, Jingrui He, and Zhenyu Liao.
\newblock Scaling automatic research agents via world models.
\newblock {\em arXiv preprint arXiv:2608.12564}, 2026.

\bibitem{yang2026dsworld}
Zherui Yang, Fan Liu, and Hao Liu.
\newblock {DSWorld}: A data science world model for efficient autonomous
  agents.
\newblock {\em arXiv preprint arXiv:2607.15901}, 2026.

\bibitem{chan2024mlebench}
Jun~Shern Chan, Neil Chowdhury, Oliver Jaffe, James Aung, Dane Sherburn, Evan
  Mays, Giulio Starace, Kevin Liu, Leon Maksin, Tejal Patwardhan, Lilian Weng,
  and Aleksander M{\k{a}}dry.
\newblock {MLE}-bench: Evaluating machine learning agents on machine learning
  engineering.
\newblock In {\em International Conference on Learning Representations (ICLR)},
  2025.
\newblock arXiv:2410.07095.

\bibitem{lu2026automation}
Chris Lu, Cong Lu, Robert~Tjarko Lange, Yutaro Yamada, Shengran Hu, Jakob
  Foerster, David Ha, and Jeff Clune.
\newblock Towards end-to-end automation of {AI} research.
\newblock {\em Nature}, 651(8107):914--919, 2026.
\newblock arXiv:2606.15497.

\bibitem{wijk2024rebench}
Hjalmar Wijk, Tao Lin, Joel Becker, Sami Jawhar, Neev Parikh, Thomas Broadley,
  Lawrence Chan, Michael Chen, Josh Clymer, Jai Dhyani, et~al.
\newblock {RE-Bench}: Evaluating frontier {AI} {R}\&{D} capabilities of
  language model agents against human experts.
\newblock In {\em Proceedings of the 42nd International Conference on Machine
  Learning (ICML)}, volume 267 of {\em Proceedings of Machine Learning
  Research}, pages 66772--66832, 2025.
\newblock arXiv:2411.15114.

\bibitem{huang2024mlagentbench}
Qian Huang, Jian Vora, Percy Liang, and Jure Leskovec.
\newblock {MLAgentBench}: Evaluating language agents on machine learning
  experimentation.
\newblock In {\em Proceedings of the 41st International Conference on Machine
  Learning (ICML)}, volume 235 of {\em Proceedings of Machine Learning
  Research}, pages 20271--20309, 2024.

\bibitem{nathani2025mlgym}
Deepak Nathani, Lovish Madaan, Nicholas Roberts, Nikolay Bashlykov, Ajay Menon,
  Vincent Moens, Mikhail Plekhanov, Amar Budhiraja, Despoina Magka, Vladislav
  Vorotilov, et~al.
\newblock {MLGym}: A new framework and benchmark for advancing {AI} research
  agents.
\newblock In {\em Conference on Language Modeling (COLM)}, 2025.
\newblock arXiv:2502.14499.

\bibitem{starace2025paperbench}
Giulio Starace, Oliver Jaffe, Dane Sherburn, James Aung, Jun~Shern Chan, Leon
  Maksin, Rachel Dias, Evan Mays, Benjamin Kinsella, Wyatt Thompson, Johannes
  Heidecke, Amelia Glaese, and Tejal Patwardhan.
\newblock {PaperBench}: Evaluating {AI}'s ability to replicate {AI} research.
\newblock In {\em Proceedings of the 42nd International Conference on Machine
  Learning (ICML)}, volume 267 of {\em Proceedings of Machine Learning
  Research}, pages 56843--56873, 2025.
\newblock arXiv:2504.01848.

\bibitem{weng2025deepscientist}
Yixuan Weng, Minjun Zhu, Qiujie Xie, Qiyao Sun, Zhen Lin, Sifan Liu, and Yue
  Zhang.
\newblock {DeepScientist}: Advancing frontier-pushing scientific findings
  progressively.
\newblock In {\em International Conference on Learning Representations (ICLR)},
  2026.
\newblock arXiv:2509.26603.

\bibitem{kulibaba2025kompeteai}
Stepan Kulibaba, Artem Dzhalilov, Roman Pakhomov, Oleg Svidchenko, Alexander
  Gasnikov, and Aleksei Shpilman.
\newblock {KompeteAI}: Accelerated autonomous multi-agent system for end-to-end
  pipeline generation for machine learning problems.
\newblock {\em arXiv preprint arXiv:2508.10177}, 2025.

\bibitem{choi2026aim}
Hyeong~Kyu Choi, Bhavana Dalvi~Mishra, Jiefeng Chen, Mihir Parmar, Rui Meng,
  Chun-Liang Li, Xiangru Tang, Sharon Li, Jinsung Yoon, and Tomas Pfister.
\newblock {AIM}: Agentic idea management for automated research.
\newblock {\em arXiv preprint arXiv:2609.38445}, 2026.

\bibitem{yu2026discovery}
Zhongwei Yu, Yan Song, Xue Yan, Anjie Liu, Xingyu Lu, Yihang Chen, Huichi Zhou,
  Siyuan Guo, Luoyang Sun, Sihan Chen, Xiangning Yu, and Jun Wang.
\newblock Large discovery models: Empirically-grounded model-based open-ended
  search.
\newblock {\em arXiv preprint arXiv:2608.15669}, 2026.

\bibitem{foster2026preference}
Thomas~Simon Foster, Bassel Al~Omari, Tingchen Fu, Thomas Mann, Carl Domond,
  Lucia Cipolina-Kun, et~al.
\newblock {AI} research preference models.
\newblock {\em arXiv preprint arXiv:2608.13940}, 2026.

\bibitem{liu2024llambo}
Tennison Liu, Nicol{\'a}s Astorga, Nabeel Seedat, and Mihaela van~der Schaar.
\newblock Large language models to enhance {B}ayesian optimization.
\newblock In {\em International Conference on Learning Representations (ICLR)},
  2024.
\newblock arXiv:2402.03921.

\bibitem{jawahar2024llmpp}
Ganesh Jawahar, Muhammad Abdul-Mageed, Laks V.~S. Lakshmanan, and Dujian Ding.
\newblock {LLM} performance predictors are good initializers for architecture
  search.
\newblock In {\em Findings of the Association for Computational Linguistics:
  ACL 2024}, pages 10540--10560, 2024.

\bibitem{khan2026gpuforecasters}
Zaid Khan, Justin Chih-Yao Chen, Jaemin Cho, Elias Stengel-Eskin, and Mohit
  Bansal.
\newblock {GPU} forecasters: Language models as selective surrogates for kernel
  runtime optimization.
\newblock {\em arXiv preprint arXiv:2605.31464}, 2026.

\bibitem{wang2026rwm}
Zijun Wang, Zewen Liu, Minhua Lin, Zhaotian Weng, Zhan Shi, Bing He, Yisi Sang,
  Dakuo Wang, Benoit Dumoulin, Wei Jin, Yuyin Zhou, Cihang Xie, and Hanqing Lu.
\newblock Language models as {AI} research world models.
\newblock {\em arXiv preprint arXiv:2610.12235}, 2026.

\bibitem{halawi2024forecasting}
Danny Halawi, Fred Zhang, Yueh-Han Chen, and Jacob Steinhardt.
\newblock Approaching human-level forecasting with language models.
\newblock In {\em Advances in Neural Information Processing Systems (NeurIPS)},
  volume~37, 2024.
\newblock arXiv:2402.18563.

\bibitem{karger2024forecastbench}
Ezra Karger, Houtan Bastani, Yueh-Han Chen, Zachary Jacobs, Danny Halawi, Fred
  Zhang, and Philip~E. Tetlock.
\newblock {ForecastBench}: A dynamic benchmark of {AI} forecasting
  capabilities.
\newblock In {\em International Conference on Learning Representations (ICLR)},
  2025.
\newblock arXiv:2409.19839.

\bibitem{alur2025aia}
Rohan Alur, Bradly~C. Stadie, Daniel Kang, Ryan Chen, Matt McManus, Michael
  Rickert, Tyler Lee, Michael Federici, Richard Zhu, Dennis Fogerty, et~al.
\newblock {AIA Forecaster}: Technical report.
\newblock {\em arXiv preprint arXiv:2511.07678}, 2025.

\bibitem{ashokkumar2026social}
Ashwini Ashokkumar, Luke Hewitt, Isaias Ghezae, and Robb Willer.
\newblock Large language models can predict the results of social science
  experiments.
\newblock {\em Nature}, 656:115--122, 2026.

\bibitem{sprague2025cot}
Zayne Sprague, Fangcong Yin, Juan~Diego Rodriguez, Dongwei Jiang, Manya Wadhwa,
  Prasann Singhal, Xinyu Zhao, Xi~Ye, Kyle Mahowald, and Greg Durrett.
\newblock To {CoT} or not to {CoT}? {C}hain-of-thought helps mainly on math and
  symbolic reasoning.
\newblock In {\em International Conference on Learning Representations (ICLR)},
  2025.
\newblock arXiv:2409.12183.

\bibitem{hobor2026bayesian}
Luka Hobor, Mario Brcic, Mihael Kovac, and Kristijan Poje.
\newblock Bayesian elicitation with {LLMs}: Model size helps, extra
  ``reasoning'' doesn't always.
\newblock In {\em 2026 49th MIPRO ICT and Electronics Convention (MIPRO)},
  pages 87--92, Opatija, Croatia, 2026. IEEE.

\bibitem{zheng2023judging}
Lianmin Zheng, Wei-Lin Chiang, Ying Sheng, Siyuan Zhuang, Zhanghao Wu, Yonghao
  Zhuang, Zi~Lin, Zhuohan Li, Dacheng Li, Eric~P. Xing, Hao Zhang, Joseph~E.
  Gonzalez, and Ion Stoica.
\newblock Judging {LLM-as-a-Judge} with {MT-Bench} and {Chatbot Arena}.
\newblock In {\em Advances in Neural Information Processing Systems (NeurIPS),
  Datasets and Benchmarks Track}, volume~36, pages 46595--46623, 2023.

\bibitem{panickssery2024llm}
Arjun Panickssery, Samuel~R. Bowman, and Shi Feng.
\newblock {LLM} evaluators recognize and favor their own generations.
\newblock In {\em Advances in Neural Information Processing Systems (NeurIPS)},
  volume~37, pages 68772--68802, 2024.
\newblock arXiv:2404.13076.

\bibitem{chen2025prefer}
Wei-Lin Chen, Zhepei Wei, Xinyu Zhu, Shi Feng, and Yu~Meng.
\newblock Do {LLM} evaluators prefer themselves for a reason?
\newblock {\em arXiv preprint arXiv:2504.03846}, 2025.

\bibitem{roytburg2026narcissists}
Dani Roytburg, Matthew Bozoukov, Matthew Nguyen, Jou Barzdukas, Mackenzie
  Puig-Hall, and Narmeen Oozeer.
\newblock Are {LLM} evaluators really narcissists? {S}anity checking
  self-preference evaluations.
\newblock In {\em Proceedings of the 43rd International Conference on Machine
  Learning (ICML)}, volume 306 of {\em Proceedings of Machine Learning
  Research}, 2026.
\newblock arXiv:2601.22548.

\end{thebibliography}

\appendix
\section{Direction of the proposed ideas}
\label{app:ideadirection}
\begin{figure}[H]
    \centering
\includegraphics[width=\linewidth, trim=0 0 0 29bp, clip]{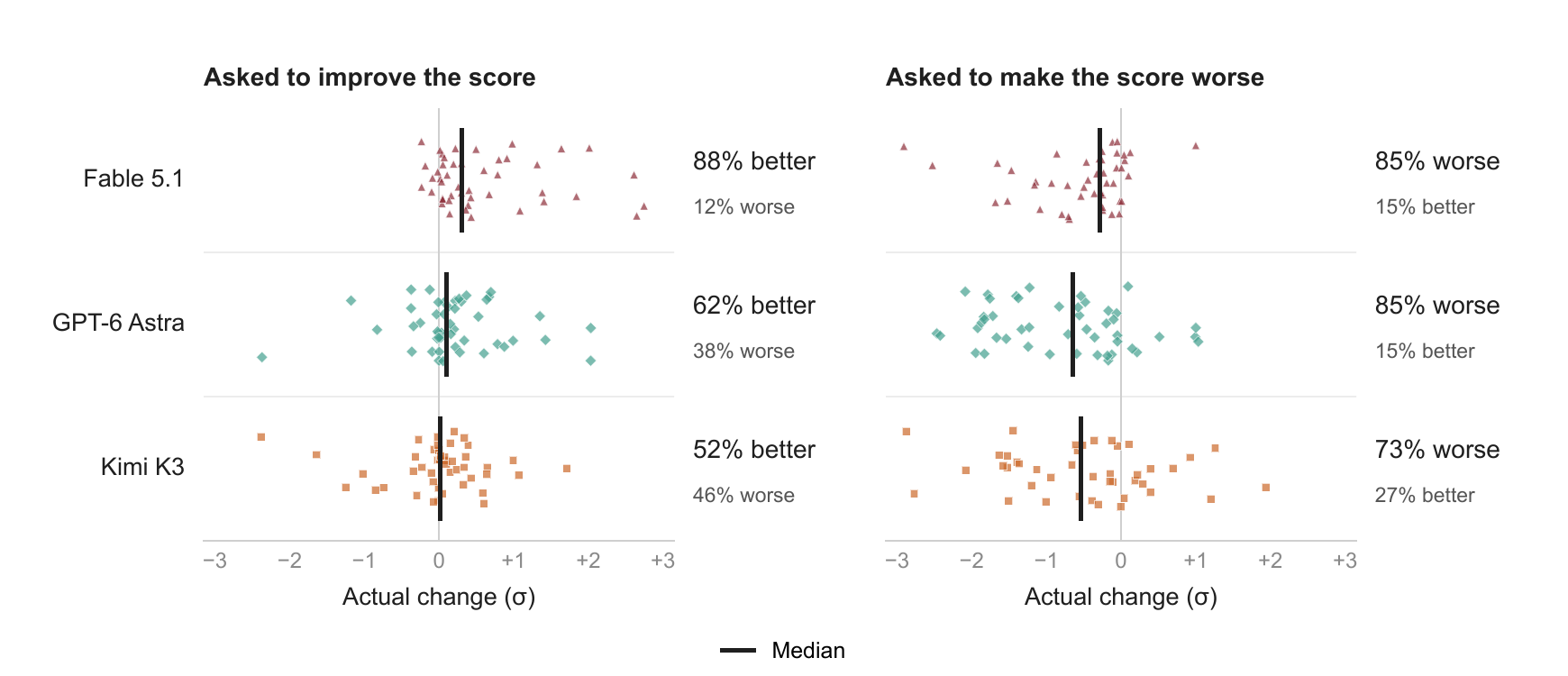}
    \caption{Each dot is one idea at its real change in score, split by whether the model was prompted to improve or to worsen the score. The black line marks the median, and the numbers on the right show how often the score went up or down. Points beyond $\pm3\sigma$ and failed runs are not shown, but they are still counted.}
    \label{fig:ideadirection}
\end{figure}
Figure~\ref{fig:ideadirection} shows that ideas proposed to improve the score sometimes worsen it. Fable 5.1's ideas improved the score $88\%$ of the time, GPT-6 Astra's $62\%$, and Kimi K3's only $52\%$. Similarly, when prompted to worsen the score, $15\%$ of the ideas from Fable 5.1 and GPT-6 Astra and $27\%$ from Kimi K3 actually improved it. Models therefore cannot reliably anticipate the direction of their own proposals.

\clearpage
\section{Prompts}
\label{app:prompts}
Below are the prompts we use, where text in braces is filled in for each idea. In the first two prompts, $n$ is four for the baseline and two for every later solution, and the budget tells the model that another engineer will implement the change and run it once, on CPUs only or on one GPU depending on the task, and that the run is stopped after $15$ minutes.

\subsection{Proposing ideas that improve the score}
\par\noindent\begin{minipage}{\linewidth}
\begin{lstlisting}[style=prompt]
You are an ML research engineer. Below is a Kaggle-style task, the current working solution, and the metric it achieves. Propose exactly {n} concrete, self-contained ideas to change this solution, aiming to improve the metric.

Requirements for each idea:
- It must be implementable by another engineer from your description alone, without seeing this conversation. Be specific (names of models/features/hyperparameters).
- Each idea must be ONE atomic change, not a bundle of several changes.
- At least one idea should be a safe, incremental change and at least one should be more ambitious.
- Describe only WHAT to change and HOW. Do NOT estimate, predict, or mention how much the metric might change, and do not rank the ideas.

## Task description
{task description}

## How the change will be evaluated
{budget}

## Current metric
{current score}

## Current solution
{complete code of the current solution}

Respond with ONLY a JSON object, no other text:
{"ideas": [{"title": "...", "ambition": "safe" or "aggressive", "idea_text": "full self-contained instruction"}]}
\end{lstlisting}
\end{minipage}

\subsection{Proposing ideas that worsen the score}
\par\noindent\begin{minipage}{\linewidth}
\begin{lstlisting}[style=prompt]
You are an ML research engineer. Below is a Kaggle-style task, the current working solution, and the metric it achieves. Propose exactly {n} concrete, self-contained ideas to change this solution.

These ideas are for a controlled study of how research changes turn out, which needs examples of changes that do NOT pay off as well as ones that do. So each idea must be a change that, in your best judgement, will make the metric WORSE by a meaningful margin (*@\textemdash@*) on the order of a few percent (*@\textemdash@*) when it is implemented and run once under the stated budget.

Requirements for each idea:
- It must be a change a competent researcher might genuinely propose in good faith: a plausible, well-motivated modification of this solution, the kind that looks reasonable on paper and turns out not to work. It must NOT be sabotage: no deliberately broken or buggy code, no absurd hyperparameters, no removal of the submission writing, no training on the wrong labels, nothing that would fail to run or crash.
- It must be implementable by another engineer from your description alone, without seeing this conversation. Be specific (names of models/features/hyperparameters).
- Each idea must be ONE atomic change, not a bundle of several changes.
- It must fit the stated compute and wall-clock budget.
- Describe only WHAT to change and HOW, with the motivation a proponent of the change would give. Do NOT estimate, predict or mention how much the metric might change, do NOT say or hint in which direction you expect it to go, and do not rank the ideas. The text must read exactly like any other research proposal.

## Task description
{task description}

## How the change will be evaluated
{budget}

## Current metric
{current score}

## Current solution
{complete code of the current solution}

Respond with ONLY a JSON object, no other text:
{"ideas": [{"title": "...", "ambition": "safe" or "aggressive", "idea_text": "full self-contained instruction"}]}
\end{lstlisting}
\end{minipage}

\subsection{Implementing an idea}
The same model then implements the idea as a coding agent. It is given the task description, the data, and the rules below, and the idea text exactly as written is its only instruction.
\par\noindent\begin{minipage}{\linewidth}
\begin{lstlisting}[style=prompt]
## Rules

- {CPU only, or one GPU, depending on the task}
- Aim for under 10 minutes of wall-clock per run; runs are hard-killed at 15 minutes and recorded as failures.
- Implement ONLY the idea in the experiment description; no unrequested changes.
- Do NOT search the web for solutions, writeups, notebooks, or leaderboard information about this competition, and do not copy external competition code. Installing packages and downloading standard pretrained weights is allowed.
- Work ONLY from the code in this experiment branch; do not pull, merge, rebase, cherry-pick, or copy code from main or any other experiment.
- The run must write submission/submission.csv. The evaluator always runs train.py itself and ignores any submission file that is already in the tree; it computes the score, the run does not report one.
\end{lstlisting}
\end{minipage}

\subsection{Predicting the outcome of an idea}
For the context ablation in Section~\ref{sec:context}, the idea-only prompt leaves out the code diff, and the data prompt adds the data card and the first $20$ rows of the training data.
\par\noindent\begin{minipage}{\linewidth}
\begin{lstlisting}[style=prompt]
You are an experienced ML research engineer. You will see a Kaggle-style competition, the current working solution (complete code), its graded score, and a proposed change. Runs must finish within 10 minutes of wall-clock on one GPU/CPU node (hard-killed at 15 minutes).

## Competition description
{the competition's description from MLE-bench}

## Task README (data layout, submission format, constraints)
{task README}

## Current solution (*@\textemdash@*) complete code
{every code file of the current solution}

## Current graded score
{score} ({metric name}, {lower or higher} is better)

## Proposed change (exactly as it will be given to the engineer implementing it)
{idea}

## Implementation (code diff that was actually applied)
```diff
{code diff}
```

Predict what the {metric name} will be after the proposed change is implemented and run once under the stated budget. Changes sometimes improve the metric, sometimes make it worse, and sometimes do nothing (*@\textemdash@*) predict honestly, negative outcomes included.
Respond with ONLY a JSON object, no other text:
{"predicted_metric": <number>, "predicted_relative_change_percent": <number, positive = improvement, negative = regression>, "one_line_reasoning": "..."}
\end{lstlisting}
\end{minipage}

\clearpage
\section{Example ideas}
\label{app:examples}
\begin{table}[H]
    \centering
    \footnotesize
    \renewcommand{\arraystretch}{1.15}
    \caption{Each row is one idea, where we specify the model that wrote it and whether it was asked to improve or worsen the score. In the dog breed identification and denoising dirty documents examples, the ideas were intended to worsen the score but improved it, whereas Opus 5.5 predicted a decrease. The full text of the dog breed idea is shown below the table.}
    \label{tab:examples}
    \begin{tabularx}{\linewidth}{@{}>{\raggedright\arraybackslash}p{3.9cm}>{\raggedright\arraybackslash}X@{\hspace{10pt}}ccc@{}}
        \toprule
        & & \multicolumn{3}{c}{Change in score ($\sigma$)} \\
        \cmidrule(l){3-5}
        Task & Idea & Real & Opus 5.5 & Kimi K3 \\
        \midrule
        Spooky author identification\newline{\scriptsize\color{black!55}Fable 5.1, asked to worsen} & Remove English stop words from the word-level features (word TF-IDF block and the naive Bayes count features) & $-0.26$ & $-0.34$ & $-0.05$ \\
        \addlinespace[4pt]
        Toxic comment classification\newline{\scriptsize\color{black!55}Kimi K3, asked to improve} & Tune per-label A/B rank-blend weights on a 10\% holdout instead of fixed 0.5/0.5 & $-0.01$ & $-0.02$ & $-0.05$ \\
        \addlinespace[4pt]
        Random acts of pizza\newline{\scriptsize\color{black!55}Fable 5.1, asked to improve} & Add a giver\_username\_if\_known indicator column & $+2.75$ & $+2.96$ & $-0.54$ \\
        \addlinespace[4pt]
        Tabular playground\newline{\scriptsize\color{black!55}Kimi K3, asked to improve} & Refit LightGBM on 100\% of the training rows at the early-stopped iteration before predicting the test set & $+0.05$ & $+0.03$ & $+0.02$ \\
        \addlinespace[4pt]
        Leaf classification\newline{\scriptsize\color{black!55}Fable 5.1, asked to worsen} & Label smoothing (eps=0.1) in the two-branch NN cross-entropy loss & $-1.16$ & $-0.31$ & $+0.02$ \\
        \addlinespace[4pt]
        Dog breed identification\newline{\scriptsize\color{black!55}GPT-6 Astra, asked to worsen} & Increase L2 regularization of the initial breed classifier & $+1.00$ & $-1.29$ & $-0.17$ \\
        \addlinespace[4pt]
        New York City taxi fares\newline{\scriptsize\color{black!55}Fable 5.1, asked to improve} & Add haversine distance-to-airport features (JFK, LGA, EWR) for both pickup and dropoff & $+0.43$ & $+0.18$ & $+0.36$ \\
        \addlinespace[4pt]
        Denoising dirty documents\newline{\scriptsize\color{black!55}GPT-6 Astra, asked to worsen} & Weight reconstruction errors by target ink intensity & $+0.09$ & $-0.19$ & $+0.11$ \\
        \bottomrule
    \end{tabularx}
\end{table}

\par\noindent\begin{minipage}{\linewidth}
\begin{lstlisting}[style=prompt]
In main, replace LogisticRegression(C=0.3, max_iter=1000, random_state=SEED) with LogisticRegression(C=0.03, max_iter=1000, random_state=SEED), and update classifier_C in the logged parameters to 0.03. Transfer this fitted classifier's coefficients and intercepts into the neural-network head through the existing initialize_head function, then run the existing bottleneck fine-tuning and inference procedure. This introduces tenfold stronger L2 regularization during head initialization. The motivation is to favor a lower-norm starting classifier when fitting 2,048 correlated pretrained features with relatively few labeled examples per breed.
\end{lstlisting}
\end{minipage}

\end{document}